\documentclass[11pt]{article}
\usepackage{scl}
\usepackage{times}
\usepackage{url}
\usepackage{latexsym}
\usepackage{lineno}
\usepackage{graphicx}
\usepackage{enumitem} 
\usepackage[toc,page]{appendix}
\usepackage{multirow}
\usepackage{longtable}

\newcommand*{\affaddr}[1]{#1} 
\newcommand*{\affmark}[1][*]{\textsuperscript{#1}}
\newcommand*{\email}[1]{\texttt{#1}}

\title{Validation and Normalization of DCS corpus using Sanskrit Heritage tools to build a tagged Gold Corpus}

\author{%
Sriram Krishnan\affmark[1], Amba Kulkarni\affmark[1], and G\'{e}rard Huet\affmark[2]\\
\affaddr{\affmark[1] Department of Sanskrit Studies, University of Hyderabad}\\
\affaddr{\affmark[2] Inria Paris Center}\\
\email{sriramk8@gmail.com, apksh.uoh@nic.in, gerard.huet@inria.fr}%
}
\date{}

\begin{document}
\maketitle
\begin{abstract}
The Digital Corpus of Sanskrit 
records around 650,000 sentences along with their morphological and lexical tagging. 
But inconsistencies in morphological analysis, and in providing crucial information like the segmented word, urges the need for standardization and validation of
        this corpus. 
Automating the validation process requires efficient analyzers which also provide the missing information. The Sanskrit Heritage Engine's Reader produces all possible segmentations with morphological and lexical analyses.
        Aligning these systems would help us in recording the linguistic differences, which can be used to update these systems to produce standardized results
and will also provide a Gold corpus tagged with complete morphological and lexical information
        along with the segmented words.
		\newcite{dataset-ak17} aligned 115,000 sentences, considering some of the linguistic differences.
As both these systems have evolved significantly, the alignment
is done again considering all the remaining linguistic differences between these systems. This paper describes the modified alignment process in detail
and records the additional linguistic differences observed.\end{abstract}

\section{Introduction}
\label{intro}

%
%
    %
    %

Computational processing of Sanskrit has been challenging due to sandhi, compounding and the free word order. Sandhi is mandatory between the components of a compound. While the sandhi between words in a sentence is left to the discretion of the speaker, due to the oral tradition, we find a greater tendency to use sandhi even in the written texts. The last decade has seen emergence of several computational tools for analysis of Sanskrit texts at various levels ranging from identification of sentence boundary \cite{hellwig-2016-detecting}, segmentation \cite{huet-morph05,hellwig-nehrdich-2018}, compound analysis \cite{anilphd}, morphological analysis \cite{kulkarni2009-IL,huet-morph05} to sentential parsing \cite{ambabook}. The complexity of the sentential parser is reduced if it receives a morphologically analysed segmented text as an input. A collaborative effort between the developers of Sanskrit Heritage (SH) Platform and Sams{\=a}dhan{\=\i} team \cite{huetkulkarni2014Coling} permitted to share efforts on these problematics. The Sanskrit Heritage Platform concentrated on the segmentation guided by the word forms validated through the lexicon. The Sams{\=a}dhan{\=\i} team focussed on the development of a parser \cite{ambabook}. The segmentation algorithm of SH Platform uses a novel approach to finite state technology, through Effective Eilenberg machines \cite{Huet-Razet15}. The non-determinism involved in segmentation as well as during the morphological analysis results in multiple possible segmentations of the given input string. The main reason behind the non-determinism is the absence of semantic compatibility check during the process of segmentation. The segmenter produces billions of possible segmentations with all relevant linguistic details such as morphological analysis, and link to the dictionary entry. In order to display these billions of solutions, an efficient compact shared representation of these solutions using tabulated display interface was developed \cite{Goyal-Huet2016}. \newcite{sk-seg:19} enlisted these solutions by ignoring the linguistic details which are irrelevant from the segmentation point of view, merging the solutions that have same word level segmentation and prioritizing the solutions with the help of statistical information from the SHMT corpus.\footnote{A corpus developed by the Sanskrit-Hindi Machine Translation (SHMT) Consortium under the funding from DeItY, Govt of India (2008-12). \url{http://sanskrit.uohyd.ac.in/scl/GOLD_DATA/tagged_data.html}}\\

Some requirements of Sanskrit computational tools are very specific. Sanskrit has a vast literature spreading over several knowledge domains. Most of the important Sanskrit literature is already translated into several languages.\footnote{Most of the literature is in poetry. In spite of having proper translations for these, the true essence of such poetry can be appreciated only in the original!} In spite of this, scholars want to have access to original sources, and thus development of the computational tools with convenient user interfaces that allow seamless connectivity to and from the lexical resources, generation engines and analysis tools becomes meaningful.
Though the user would like to have an access to all possible interpretations, it is desirable to rank the solutions and display a few of them. Only when none of the displayed solutions is correct, all other solutions should be made available to the user. In order to rank the solutions, one needs some annotated corpus which can be used to learn the priorities.\\


For this, Heritage segmenter is required to be facilitated with statistical analysis.
Recently the digitization of Sanskrit manuscripts shot up. But the amount of annotated data available for Sanskrit is very small compared to the size of the texts available in it from ancient times. 
An effort towards having such an annotated data was initiated and resulted into the Digital Corpus of Sanskrit (DCS) \cite{dcs-oh}.\footnote{\url{http://www.sanskrit-linguistics.org/dcs/}} This data, being of reasonable size, can be used for both statistical analyses and use of machine learning algorithms.\\

This paper focuses on how DCS's data can be used along with the Heritage Engine's analysis so that we get a proper morphologically tagged and segmented
corpus. It starts with describing the annotation schemes of the two systems, their advantages and limitations and the need for alignment. A similar effort towards aligning the DCS annotated data with the analysis of Heritage Engine's data was already reported by \newcite{dataset-ak17}. This work is briefly described in section \ref{dataset}, along with 
some issues related to the alignment. Looking at the limitations of the previous work, an effort towards building a better dataset was started. A proper alignment between the representations of these systems was done. The alignment process is described in section \ref{align}. But there were some additional difficulties due to the differences in the design decisions of the two systems. These difficulties were recorded systematically and are discussed in section \ref{prob}. The observations are put down in section \ref{obs}.\\



\section{Resources}

Before going into the discussion, let us take a look at the morphological annotation and the segmentation annotation of The Digital Corpus of Sanskrit, and The Sanskrit Heritage Engine.

\subsection{The Digital Corpus of Sanskrit}
\label{dcs}

The DCS consists of a Sanskrit corpus in 650,000 text lines. It is a sandhi split corpus of Sanskrit texts with full morphological and lexical analysis. There are more than 4,500,000 word references with around 175,000 unique words. All this data was collected from around 400 Sanskrit texts. \newcite{dataset-ak17} represented this data as objects containing the sentence details like chunks, lemmas, and CNG values.\footnote{A value denoting the case, number, and gender of the given word, for nouns. Or the tense, aspect, person, number, and ga\d{n}a for verbs.} A glimpse of what an object looks like is depicted in Table \ref{tab:pickle}. This object was
used for alignment with the analysis done by Heritage Engine.\\

\begin{table}[h]
\begin{center}
\scalebox{0.9}{
\begin{tabular}{|l|l|}\hline
Sentence Id & 83\\\hline
Sentence & mauktike yadi sa\d{m}deha\d{h} k\d{r}trime sahaje'pi v\={a}\\\hline
Chunks & [`mauktika', `yadi', `sa\d{m}deha', `k\d{r}trima', `sahaja', `api', `v\={a}']\\\hline
Lemmas & [[`mauktika'], [`yadi'], [`sa\d{m}deha'], [`k\d{r}trima'], [`sahaja'], [`api'], [`v\={a}']]\\\hline
Morphological Class (CNG) & [[`171'], [`2'], [`29'], [`171'], [`171'], [`2'], [`2']]\\\hline
\end{tabular}}
\caption{An example DCS Object data}
\label{tab:pickle}
\end{center}
\end{table}
DCS presents a lemma for each segment along with its morphological analysis.
In some cases the derived stem is chosen as a lemma, and in some cases the underived one is chosen. The system is not uniform in deciding the lemmas. This might be a result of context-specific analysis, but in the absence of any tagging guidelines, we do not know the reason for such inconsistency.
This corpus is curated single-handedly. Thus we can assume consistency in tagging. However, every human is prone to error. The quality of this data tested on a small sample \cite{hellwig-nehrdich-2018} revealed that around 5.5\% of the compound splits are doubtful and around 2\% errors are due to segmentation.
Recently released DCS data \cite{hellwig-nehrdich-2018}, contains the split points and sandhi rules proposed by the tagger.

\subsection{The Sanskrit Heritage Engine}
\label{she}

The Sanskrit Heritage Engine is a platform that hosts a lexicon (The Sanskrit Heritage Dictionary) and various tools like reader, lemmatizer, declension and conjugation engines. The Reader analyses any given text and segments it into all possible splits and displays them in a graphical interface where the user has the option to choose the required split. In addition to the graphical interface, it also enlists the solutions if the number of solutions is less than a threshold, say 100, also providing the sentential analysis. 
It closely follows P\={a}\d{n}ini's system i.e., all the rules governing the concept of \textit{sandhi }that occur in \textit{A\d{s}\d{t}\={a}dhy\={a}y\={i}} are taken into consideration. This segmentation is lexicon directed, using forms systematically generated from its own lexicon.\\

Additionally, the morphological analyses (inflectional and if applicable derivational as well) are also provided for all the segments. This helps the user to disambiguate and correctly pick the intended split and prune the solutions that are not required. Such information also helps in the further stages of sentential analysis like parsing, disambiguation, and discourse analysis. Another advantage is that, this system combines a fast segmentation algorithm using finite-state transducers and dynamic programming with a first-pass of chunking that limits the inherently exponential complexity to small-length chunks, making the whole segmentation analysis fast enough in practice to be usable interactively.\\

One limitation of this system is that, it cannot arrive at a single solution mechanically. 
This owes to the fact that the current version does not take into account the meaning compatibility between various segments, which involves processing at sentential level. Another limitation is due to the Out of Vocabulary words. Though the dictionary contains high frequency words, as is the case with any NLP system, Heritage engine also suffers from automatic handling of Out of Vocabulary words. However the interactive interface allows the user to suggest the lemma for such words, which get stored in the local dictionary. This is done for nominal words, and if the lemmas are available in the Monier-Williams dictionary. And for some nominal words with certain prefixes like \textit{su, vi, dur} and for words with \textit{taddhita} suffixes, an explicit entry is required to be added in the dictionary.\\
This paper deals with the ways in which the DCS GOLD data is aligned with one of the solutions of SH. It further provides the detailed analyses of the difficulties encountered during the alignment. This is an extension of the effort by \newcite{dataset-ak17} which is described in the next section.

\section{A Dataset for Sanskrit Word Segmentation}
\label{dataset}

\subsection{Description}

Around 115,000 sentences from the originial DCS corpus were considered and an alignment between the DCS and the Heritage Engine was done by \cite{dataset-ak17}. This lead to a huge dataset which had the input sequence, ground truth segmentation, and morphological and lexical information about all the phonetically possible segments. This was done primarily for the Word Segmentation task but is also useful for subsequent tasks. The main concern was to make this annotated corpus available for the use of statistical techniques and Machine learning algorithms.\\

Since there are differences in design decisions between DCS and Heritage Reader, the candidate segments provided by the Heritage Reader had to be adapted and a few additional segments were added so as to match the entries in DCS. The XML based GraphML format was used to represent the candidate space segments. The GraphML files consist of graph structures, $G(V,E)$ as the representation for the analyses of each of those sentences. The nodes, $V$,  are the possible splits, and the values in the edges, $E$, denoting whether the participating nodes can co-exist in a solution or not i.e., whether or not they have an overlap in the position relative to the sentence, and that the overlapped portion does not follow any sandhi rule.\\

To match the two systems, the data from the Heritage Reader's analysis was scrapped and certain parameters such as word, lemma, position, morphological information, chunk number, word length, and pre-verbs were extracted.  Corresponding to the morphological analysis the CNG was generated for the ease of alignment. With all these parameters as attributes of each of the nodes, graphs were build for each sentence. Standard graph processing libraries were used to extract the data from these graphs.

\subsection{Issues already handled}
The lemma provided by DCS and the CNG value are the attributes that help in mapping the two systems. Although direct mapping produced some results, there were multiple issues when the actual mapping was experimented with. These issues and their solutions were discussed in \cite{dataset-ak17}. We provide a short summary of these issues here.
\begin{itemize}
	\item One of the issues was with the compounds and Named entities. The DCS provided the lemma based on the context. If in a given context the compound has non-compositional meaning, then the compound was not split into its components. If the meaning in the given context is compositional, then the compound was split into components. The Heritage Reader's analysis is guided by the lexicon. If the lexicon contains a compound entry due to its non-compositional meaning or on account of being a Named entity, then the Reader in addition to producing all possible segmentations, also produced an entry without any segmentation. Deciding non-compositionality is a tricky issue \cite{hellwig-nehrdich-2018}.

\item Another issue was with the derivative affixes, especially the secondary derivatives. These were treated separately in Heritage Reader, but DCS joins them.

\item Third, there were inconsistencies while dealing with \textit{anusv\={a}ras} which should have actually been the \textit{anun\={a}sikas} when followed by their respective varga consonants.

\item Fourth, in some cases the difference in DCS was with certain compounds where the \textit{p\={u}rva-pada's} lemmas were same as their compounding forms. But the Heritage Reader sticks to the P\={a}\d{n}inian rules and produces only the base stem. 

\item Finally, DCS had its lemmas joined with the pre-verbs, but the Heritage Reader displays them separately. 
\end{itemize}

\subsection{Issues yet to be handled}
\label{issue}

In addition to the above mentioned differences, we noticed following discrepancy at the level of analysis between the two systems.

\begin{itemize}
\item Use of homonymy index\\
	One important aspect of the Heritage Reader is that the dictionary has different entries for homonymous stems, and the morphological analyser provides the homonymy index of the stem. For example, the word \textit{siddham} is analysed by Heritage Engine as shown below.
\begin{quote}
	\begin{verbatim}
	[siddha_1 { pp. }[sidh_1]]{n. sg. acc.| n. sg. nom. | m. sg. acc}
	[siddha_2 { pp. }[sidh_2]]{n. sg. acc.| n. sg. nom. | m. sg. acc}
	\end{verbatim}
\end{quote}

If we look at the meanings of these two senses, we find that they are almost opposing each other.

\begin{quote}
	\begin{verbatim}
	siddha_1  [pp. sidh_1] a. m. n. f. siddh\={a} 
	(French)  accompli, réalisé ; gagné, obtenu ; parfait
	          qui a atteint son but, réalisé son objectif
	(English) accomplished, realized; won, obtained; perfect
	          who achieved his goal, achieved his goal
	siddha_2  [pp. sidh_2] a. m. n. f. siddh\={a}
	(French)  empêché, écarté, repoussé.
	(English) prevented, pushed aside, pushed back
	\end{verbatim}
\end{quote}
		
In DCS annotated data we have not come across any homonymy index and hence \newcite{dataset-ak17} collapsed these two analyses into one ignoring the sense information. Although it is not used now, such distinctions would definitely be of greater use for sense disambiguation of such homonymous words. 

\item Level of analysis\\
The engine provides both the inflectional as well as the derivational analysis for some \textit{k\d{r}dantas} (primary derivatives) i.e, participles, absolutives and infinitive forms. For example, the word \textit{hitam} is an inflected form of the derived root \textit{hita} which can be derived in two different ways. This results in two different analyses for the word \textit{hitam} as shown below.
\begin{quote}
\begin{verbatim}
[hita_1 { pp. }[hi_2]]{n. sg. acc. | n. sg. nom. | m. sg. acc.}
[hita_2 { pp. }[dh\={a}_1]]{n. sg. acc. | n. sg. nom. | m. sg. acc.}
\end{verbatim}
\end{quote}

The dictionary entries for these two stems show the meaning difference. 
\begin{quote}
\begin{verbatim}
hita_1    [pp. hi_2] a. m. n. f. hita
(French)  envoyé, lancé, émis.
(English) sent, launched, issued
hita_2    [[pp. dh\={a}_1] a. m. n. f. hita
(French)  placé, mis, disposé | convenable, avantageux ;
          utile, propre à, bon pour <dat. g. loc.> ; 
          salutaire | amical, bienveillant ; qui fait le bien
          avantage, profit, intérêt ; bien, chose utile ; bien-être.
(English) placed, put, disposed | suitable, advantageous;
          useful, suitable for, good for <dat. g. loc.>;
          beneficial | friendly, caring; who does good
          advantage, profit, interest; well, useful thing; well-being.
\end{verbatim}
\end{quote}

As already pointed out, we noticed that there is no uniformity in the analysis in DCS data. Sometimes the derived words are analysed providing the derivational analyses and sometimes they are not analysed. 

For example, in the case of \textit{hitam}, DCS chooses only the inflectional analysis. In the case of causative forms of the verbs, DCS chooses the causative form of the verb \textit{bhojay} as the stem for the word \textit{bhojan{\=\i}y{\=a}\d{h}}. But the SH provides the following analysis for the same word.\\
\begin{quote}
\begin{verbatim}
[bhojan\={i}ya {pfp. [2] }[bhuj_2]]{f.pl.acc. | f.pl.nom | m.pl.nom.}
[bhojan\={i}ya {ca. pfp. [2] }[bhuj_2]]{f.pl.acc. | f.pl.nom | m.pl.nom.}
[bhojan\={i}ya {pfp. [2] }[bhuj_1]]{f.pl.acc. | f.pl.nom | m.pl.nom.}
\end{verbatim}
\end{quote}

		Thus here one needs to construct the causative form \textit{bhojay} from \textit{bhuj}+\textit{ca} in order to align the morphological analysis, which is not trivial and involves the rules from grammar. Additionally, the Heritage Engine analyses privative compounds like \textit{aniv\d{r}ttam} as \textit{a-niv\d{r}tta}, but DCS treats such compounds to have non-compositional analyses.\\

\item Enhancement in the Heritage engine\\
The Heritage Reader's Engine and the dictionaries have evolved in many ways in the past three years, and hence using the same GraphML files would neglect the improvements carried over during the last few years. One such change was with the way the Named Entities have been handled. In the earlier version, the compounds were always split into possible segments even if it represents a Named Entity. 
Another modification was regarding the pre-verbs. Earlier only the pre-verbs with derivational lemmas were joined, but in the current version 
the inflectional lemmas also have the pre-verbs attached alongwith. \\


\end{itemize}
In view of the above changes, and in order to resolve the problems with homonymy, we decided to align the Heritage Reader's analysis with the DCS afresh with modifications in the alignment process.
This alignment of the manually tagged analyses of DCS with one of the analyses produced by the Heritage Engine would provide us with:
\begin{enumerate}[nosep]
	\item Identifying wrong annotations from DCS,
	\item Consistent uniform analysis,
	\item Probable compounds with non-compositional meaning, and
	\item Constituency analysis of compound words
\end{enumerate}

\section{Alignment Process}
\label{align}

The DCS objects for the sentences are used in the same way they were used earlier. GraphML files are used for the representation of the mapped data. The same process of scrapping was used with a slight modification to scrap the derivational information, sense, and the correct representation of the morphological analysis. 


The Mapping for every sentence was done in three stages:

\begin{enumerate}[nosep]
	\item Representing Heritage Reader analysis as a graph,
	\item Aligning the DCS annotation with Heritage Reader's analysis, and
	\item Handling compounds with non-compositional meaning and words with derivational morphology
\end{enumerate}

\subsection{GraphML files creation}

\subsubsection{Scrapping}

The first stage corresponds to scrapping data from the Heritage Reader's website and creation of GraphML files for each of the sentences from the DCS corpus. During this stage, CNG values corresponding to each morph analysis is also obtained, and added as an additional attribute to help in connecting the Heritage Reader's analyses with the DCS entries \cite{dataset-ak17}.


\subsubsection{Graph Construction}

The next part of this stage was creating graphs with nodes having the following form.

\begin{quote}
\begin{verbatim}
( id, { color_class, position, chunk_no, word, lemma, sense, cng, pre_verb, 
morph, length_word, der_pre_verb, der_lemma, der_sense, der_morph, der_cng, 
char_pos } )
\end{verbatim}
\end{quote}

All these values are extracted from the scrapped data. lemma, sense, pre-verb, morph and cng denote respectively the word's pr\={a}tipadika/dh\={a}tu (stem/root), sense (based on different meanings), upasarga (pre-fix for verbs), morphological details, CNG - (case number and gender value) corresponding to the morphological information.
der\_lemma, der\_sense, der\_pre\_verb, der\_morph, der\_cng correspond to the information pertaining to derivational morphology.

Except for the derivational details, the sense information and the position based on character, all the others were created by \newcite{dataset-ak17}. As in \newcite{dataset-ak17}, the edges are added between the nodes with values as either `1' or `2', based on the positions of the words in the participating nodes considering the position overlap in the sentence. If there exists a conflict, `2' is added, else `1'. For example, in the compound \textit{p\={a}t\={a}labh\={a}suram}, \textit{p\={a}t\={a}la} and \textit{bh\={a}suram} are non-conflicting nodes and hence their edge is labeled `1'. But \textit{bh\={a}} and \textit{bh\={a}suram} are separate nodes which are conflicting, and hence their edge is labeled `2'.

As another example, let us consider the sentence \textit{bindusth\={a}na\d{m} madhyade\'{s}e sad\={a} padmavir\={a}jitam}. There are 2 possible ways in which the word \textit{padmavir\={a}jitam} can be analysed as depicted in Table \ref{tab:pdmvrj}.
So, the word \textit{vir\={a}jitam} is in conflict with the part \textit{vir\={a}ji} since they cannot co-occur together, and hence they will have an edge labeled as `2'. The nodes with \textit{vir\={a}ji} and \textit{padma} will have an edge with label `1' since they can co-occur. Currently the edge information is not used but could be of use later.

\begin{table}[h]
\begin{center}
\begin{tabular}{|l|l|}\hline
Solution & Analyses\\\hline
\multirow{2}{*}{padma-vir\={a}jitam} & padma [padma]\{iic.\}\\
& vir\={a}jitam [vi-r\={a}jita \{ pp. \}[vi-r\={a}j\_1]]\{n.sg.acc. | n.sg.nom. | m.sg.acc.\}\\\hline
\multirow{3}{*}{padma-vir\={a}ji-tam} & padma [padma]\{iic.\}\\
& vir\={a}ji [vir\={a}j\_2]\{m.sg.loc. | n.sg.loc. | f.sg.loc.\}\\
& tam [tad]\{m.sg.acc.\}\\\hline
\end{tabular}
\caption{Analysis of the chunk padmavir\={a}jitam}
\label{tab:pdmvrj}
\end{center}
\end{table}

\subsubsection{Handling Homonymy}

This involves merging the nodes with same lemma and other parameters (like word, chunk, etc.) but different senses. Since the DCS does not differentiate lemmas based on their senses, 
the individual nodes, having different senses, but the same lemma and same CNG value lead to multiple mappings with the DCS.
Such nodes are collapsed into one, suppressing the information of senses. It was mentioned in section \ref{issue} that \newcite{dataset-ak17} collapsed all these nodes into one. Since, we have an additional attribute in `sense', all the sense indices are temporarily stored in this attribute. New graphs were formed with the new nodes.

%
%
%

\subsection{Comparison of DCS data and Heritage Reader's graph to analyze the parallels}

The second stage of the alignment process deals with the actual analysis for creating the merged parallel database. First, both systems are normalized based on the \textit{anun\={a}sika} rules so that both of them have a uniform representation of the texts. Then the mapping process is initiated.

\subsubsection{Normalization}

Both the DCS data as well as the Heritage output is normalised to account for the variations in the use of anun{\=a}sika and doubling of consonants (\textit{dvitva}).
\begin{quote}
\textit{\'{s}r\={i}\'{s}a\d{m}kara\d{h}} to \textit{\'{s}r\={i}\'{s}a\.{n}kara\d{h}}
\end{quote}

\subsubsection{DCS and Heritage Reader comparison}

A comparison between the modified DCS data and modified Heritage Engine analyses was done sequentially with the following for matching:
\begin{enumerate}
	\item Mapping the lemma and CNG.
	\item Mapping the derived stem of Heritage Engine with the DCS's lemma.\\
	For example, the word \textit{kartavy\={a}} is analysed by Heritage Reader as 
	\begin{quote}
		[kartavya { pfp. [3] }[k\d{r}\_1]]{f. sg. nom.}
	\end{quote} and the lemma in DCS is `\textit{k\d{r}}'. So, the derivational lemma \textit{k\d{r}\_1} is taken into account in this case.
	\item Different conventions\\
		In the case of pronouns, both DCS and Heritage engine follow different conventions regarding the stem.
		For example, DCS  analyses \textit{tvam} as \textit{tvad}, and the Heritage output following P\={a}\d{n}ini produces \textit{yu\d{s}mad} as the stem. 
		These are treated as special cases with normal table lookup.
	\item Mapping compound iics\footnote{\textit{in initio compositi}}\\
		In the case of iics, DCS does not provide the lemma but just provides the split point in a compound.
		For example, \textit{mah\={a}deva} has the lemmas in analyses as \textit{mah\={a}-deva} in DCS, but \textit{mahat-deva} in Heritage Reader. In such cases we align the segments taking into consideration the segment and not the lemma.
\end{enumerate}

It was then observed that there were mappings with exactly one match, more than one match, at least one lemma not matched, and both multiple matches and missed matches present for a single sentence.

So the results are categorized into 4 groups:
\begin{enumerate}
	\item Single parallel mapping obtained for all lemmas in the sentence
    \item Sentences that have at least one lemma with multiple parallels
    \item Sentences that have at least one lemma without any parallel
    \item Sentences that have at least one lemma with multiple parallels and at least one lemma without parallels
\end{enumerate}
    
The unmapped sentences (3 and 4) are then sent for further modifications.

\subsection{Modifications}

Three kinds of modifications are done:
\begin{itemize}
	\item Direct mapping of lemmas of verbs in the tenth ga\d{n}a, having causative suffix \textit{\d{n}ic}.

        Eg, \textit{p\={u}jayati} is analysed in Heritage Engine as `\textit{[p\={u}j]{pr. [10] ac. sg. 3}}', but DCS has the lemma as \textit{p\={u}jay}. A separate list of such pairs like \textit{p\={u}jay-p\={u}j}, \textit{bh\={u}\'{s}ay-bh\={u}\'{s}}, etc, was prepared. Additional nodes are created using such matching entries.

	\item The preverbs are sandhied with their corresponding lemmas, and derivational lemma is sandhied with its corresponding preverb labeled as der\_pre\_verb

        Eg, \textit{pra\'{s}a\d{m}santi} is analysed as `\textit{[pra-\'{s}a\d{m}s]{pr. [1] ac. pl. 3}}' in Heritage Reader, but DCS has the lemma \textit{pra\'{s}a\d{m}s}. In this case, a new node with lemma \textit{pra\'{s}a\d{m}s} is created by performing sandhi between the preverb \textit{pra} and lemma \textit{\'{s}a\d{m}s}. Care should be taken when such sandhi is done, since there are certain required tranformations such as retroflexion of n (\d{n}) and s (\d{s}). For example, \textit{pra} and \textit{nam} become \textit{pra\d{n}am}. Finally these new nodes are added to the graph.

	\item Merged possible components of compounds to form individual lemmas

		Eg, \textit{\'{s}a\.{n}kha\'{s}uktyudbhavam} is analysed in Heritage Reader separately as \textit{\'{s}a\.{n}kha-\'{s}ukti-udbhavam}, but \textit{\'{s}a\.{n}kha-\'{s}uktyudbhavam} is the expected solution according to DCS. So, for this chunk, all possible compounds are constructed and then each of it is compared to the DCS analysis. If the correct one is matched, a new node with the modified lemma, word and other information is created in the graph. The value for the attribute word, is kept as a hyphen-separated compound instead of the sandhied compound for future usage in constituency analysis. We should also make a note here that the total number of combinations is a Catalan number. So, generating all possible combinations for a given compound results in exponentially slow algorithm.
\end{itemize}

After these modifications, the second stage of analyzing the parallels is done for the modified graphs. Together with the previous results, the number of sentences with single parallels and multiple parallels is taken into consideration for observations.

\section{Additional Problems} \label{prob}

After the modifications, the mapping resulted in a reasonable amount of success. But there were still issues regarding the lemmas that didn't match at all, and the lemmas that had multiple mappings. Let us first look into some of the issues that lead to the lemmas to have multiple mappings.

Mapping for the CNG values is not one-to-one. For a given CNG value, there could be multiple morphological analyses. For example, the CNG value of -190 has morphological analyses as `ca. pp.', `des. pp.', and `pp.'. The DCS doesn't distinguish between these analyses. Hence, we find multiple analyses being mapped for the same lemma. For the sentence \textit{vasur \={a}dya\d{m} \'{s}iva\d{m} c\={a}dya\d{m} m\={a}y\={a}binduvibh\={u}\d{s}itam}, the lemma that had multiple parallels was \textit{vibh\={u}\d{s}ay}. On observing the morphological information, the difference between the multiple solutions was in the morphological analyses of the Heritage Engine. There were two entries with the same lemma - \textit{vibh\={u}\d{s}ay}. One had the morph as `pp.' and the other had it as `ca. pp.' (causative pp.). The DCS clubs both of them and assigns the CNG value as -190.

The DCS either considers only the derivational CNG value, or the inflectional CNG value, and not both of them together. When it considers the derivational CNG, it does not give any information about the CNG of the inflected form. In such a case, there could be multiple possibilities since the inflectional forms can be ambiguous. For the sentence \textit{\'{s}ruta\d{m} vede pur\={a}\d{n}e ca tava vaktre sure\'{s}vara}, the lemma \textit{\'{s}ru} has CNG value -190 according to DCS. The Heritage Reader has the derivational lemma as \textit{\'{s}ru}, and hence derivational CNG as -190. But the inflectional lemma formed is \textit{\'{s}ruta} and there are three different entries with three different CNG values (31, 71, 69) and hence multiple parallels were obtained.

The DCS is also not uniform in assigning the lemma. Sometimes, even when the derivational lemma is available, it uses the inflectional lemma. For example, in the sentence \textit{n\={i}la\d{m} n\={i}la\d{m} sam\={a}khy\={a}ta\d{m} marakata\d{m} harita\d{m} hitam}, the analysis for \textit{hitam} is provided as the inflectional form \textit{hita}, and not it's derivational form \textit{hi} or \textit{dh\={a}}.
So, it is hard to map them because there is some amount of information missing. To handle these, the DCS entries are to be modified, which is a huge and tedious task.

Let us now look at an issue with compounds. For the sentence \textit{ata eva hi tatr\={a}dau \'{s}\={a}nti\d{m} kury\={a}d dvijottama\d{h}}, the compound \textit{dvijottama\d{h}} has its lemma as \textit{dvijottama} in the DCS. But the compound modifications done to Heritage reader's solutions leads to multiple entries with the same lemma. There are four possibilities: \textit{dvija-uttama\d{h}}, \textit{dvi-ja-uttama\d{h}}, \textit{dvij\={a} uttama\d{h}}, and \textit{dvi-j\={a} uttama\d{h}}. Of these, the last two are analysed as two words, and not a compound, and hence neglected but the first and the second are both possible compounds. In many such compounds, the differences are due to the combinations of the components. 

In the Heritage's analysis for the sentence \textit{vada me parame\'{s}\={a}na homaku\d{n}\d{d}a\d{m} tu k\={i}d\d{r}\'{s}am}, the word \textit{parame\'{s}\={a}na} is analysed as \textit{parama-\={i}\'{s}\={a}na}. There are three analyses for the word \textit{\={i}\'{s}\={a}na}. Two have participial forms generated from the root \textit{\={i}\'{s}}, and the other is generated from the lexicon entry \textit{\={i}\'{s}\={a}na [agt. \={i}\'{s}\_1]} which states that it is an agent noun of the root \textit{\={i}\'{s}}. The participial forms and the agent noun belong to different phases in the analysis - Krid and Noun, respectively. This distinction is present in the Heritage Reader because pre-verbs are not attached to general nominal entries but attached to participles. So, the nominal entry needs to be treated separately, and hence Noun is more preferred than being analysed as derived from the root \textit{\={i}\'{s}}. Since DCS doesn't have this distinction, we arrive at multiple alignment. But, in this case, the agent noun, being derived from \textit{\={i}\'{s}}, is indeed a \textit{k\d{r}danta} and should be present alongwith the other two.

In the sentence \textit{k\={a}ra\d{n}ena mah\={a}mok\d{s}a\d{m} nirm\={a}lyena \'{s}ivasya ca}, the word \textit{mah\={a}mok\d{s}a\d{m}}'s analysis, according to DCS, has its lemma as \textit{mah\={a}mok\d{s}a} with the CNG as 31. Such compounds' contexts need to be checked whether they are to be treated as compounds with non-compositional meaning, or whether they are to be split further.

Now, looking at the unmatched lemmas, we find the following difficulties. Some words are not analysed by the Engine at all either due to absence of the word and its \textit{pr\={a}tipadika} in the dictionary or, the engine fails to analyse the words. For example, the word \textit{prameyatvam} is analysed by DCS as having the lemma \textit{prameya} and \textit{tvam}, but the Heritage Engine produces only parasite segmentations, in the absence of a lexical entry for \textit{prameyatva}. Similarly certain \textit{taddhit\={a}ntas} (secondary derivatives) like \textit{nirgu\d{n}atvam} are not analysed in the Heritage Engine. Further modifications to the engine to analyse the secondary derivatives would help in aligning these words.

The secondary derivatives are treated like compounds in DCS. For example, in the sentence \textit{prakury\={a}t tu dvijenaiva tad\={a} brahmamay\={i} sur\={a}}, the word \textit{brahmamay\={i}} is analysed as a compound of \textit{brahman} and \textit{maya}. The Heritage engine analyses it with the lemma as \textit{brahmamaya}. Since DCS does not differentiate between compounds and words with secondary derivative suffixes, it is not possible to align such words with the Heritage Engine's analysis.

There is another issue with the way indeclinables like \textit{api} are handled. The Heritage Reader analyses it as `conj.' and `prep.'. But DCS marks it as `ind.' and assigns the CNG as 2. A normalization is required to classify properly such indeclinables under `conj.', `prep.' etc.

Sometimes the distinctions are present in the way certain \textit{p\={u}rva-padas} of compounds are analysed. For example, the component \textit{r\={u}pya} has the analysis as \textit{r\={u}pya} with the morphological analysis as `iic'. But the Heritage Reader provides the lemma as \textit{r\={u}pya} with morphological analysis as `pfp. iic.'.

These are just a handful of examples of the issues encountered. Further analysis of the missed alignments will bring out more such difficulties, but will also provide an opportunity to modify the corpus and the two systems.

\section{Observations} \label{obs}

In total, 119,502 sentences were checked for the parallels. 92,781 sentences were successful in finding matches for all their lemmas. Out of these, 39,793 sentences had exactly one match. For the remaining 52,988 sentences, there were multiple mappings for at least one lemma in each of those sentences. The word and transition frequencies were also obtained. Of the 67,330 sentences sent for modifications, 12,639 sentences were not modified at all. So, they had at least one lemma not mapped. The remaining 14,082 sentences, although were modified, had at least one lemma not mapped. So, they have to be considered individually for understanding the reasons for not being mapped.

\section{Conclusion}

This dataset provided around 77\% mappings with around 42\% of which had single mapping. The previous dataset ended in more mappings comparatively, but did not consider the linguistic details, this paper focuses on. These issues are definitely needed to be considered to make the dataset devoid of errors. In further releases, the number of missed lemmas and multiple parallels are to be reduced. Concentrating on the issues recorded would give us insights on how to proceed further for building a better annotated corpus.

\nocite{*}

\bibliographystyle{acl}
\bibliography{dcs_she}

\end{document}